\documentclass[11pt]{article}

\usepackage[margin=1.15in]{geometry}
\usepackage{booktabs}
\usepackage{array}
\usepackage{amsmath}
\usepackage{graphicx}
\usepackage[hidelinks]{hyperref}
\usepackage{authblk}
\usepackage{caption}
\usepackage[numbers,sort&compress]{natbib}

\title{\vspace{-1.5em}Prompting is not enough: supervised baselines and
leakage control for measuring shared decision-making with LLMs in
pediatric encounters}

\author[1,4]{Bernardo Modenesi}
\author[2]{Jody L. Lin}
\author[3]{Kimberly A. Kaphingst}
\author[2]{Angela Zhu}
\author[2]{Maya Wheeler}
\author[1]{Peilu Zhang}
\author[1]{Angela Fagerlin}
\affil[1]{Department of Population Health Sciences, University of Utah
Spencer Fox Eccles School of Medicine, Salt Lake City, UT, USA}
\affil[2]{Division of Hospital Medicine, Department of Pediatrics,
University of Utah Spencer Fox Eccles School of Medicine, Salt Lake City,
UT, USA}
\affil[4]{Kahlert School of Computing, University of Utah, Salt Lake City,
UT, USA}
\affil[3]{Department of Communication and Huntsman Cancer Institute,
University of Utah, Salt Lake City, UT, USA}
\date{}

\begin{document}

\maketitle
\vspace{-1.5em}

\section*{Abstract}

\textbf{Objectives.} To determine whether zero-shot prompting of a large
language model (LLM) is sufficient to detect shared decision-making (SDM)
behaviors in real clinical encounters and whether supervised learning adds
value under patient-grouped, nested evaluation.

\textbf{Materials and Methods.} We analyzed 21 audio-recorded outpatient
surgical decision encounters (19 unique patients; 7{,}566 utterance
segments; approximately 6.1 hours) between families of children with
multiple long-term conditions and their surgical providers. Trained coders
labeled segments for 12 SDM behaviors (human--human macro Cohen's
$\kappa = 0.695$). We compared a zero-shot local LLM (Qwen 2.5 32B), a
supervised classifier over frozen sentence embeddings, and their logistic
stack under patient-grouped outer folds. Model fitting and threshold selection
used inner patient-grouped predictions, and confidence intervals resampled
patients.

\textbf{Results.} The zero-shot LLM reached macro $\kappa = 0.139$ (95\% CI
0.111--0.164). The supervised classifier reached $\kappa = 0.227$
(0.186--0.262), a paired improvement of $\Delta\kappa = 0.088$
(0.051--0.119). A logistic stack of the zero-shot and supervised scores
reached $\kappa = 0.242$ (0.198--0.284). We identified multiple
corpus-specific leakage paths, including grouping sibling recordings
separately and allowing labels from an outer held-out patient to enter
few-shot exemplars used while fitting downstream models.

\textbf{Discussion.} In this corpus and model configuration, zero-shot
prompting underperformed a modest supervised baseline. Patient-level grouping
alone was insufficient to prevent leakage when labeled prompt exemplars were
precomputed outside the outer evaluation loop.

\textbf{Conclusion.} In this single-site corpus, zero-shot prompting alone
is not sufficient to measure SDM behavior as reliably as a small supervised
model. Reported performance is sensitive to the unit of data splitting and to
where labeled prompt exemplars enter the evaluation pipeline. External
validation is needed before these findings generalize beyond this population,
model, prompt, and codebook.

\section{Background and Significance}

Children with medical complexity (CMC) make up fewer than 5\% of pediatric
patients but account for over 30\% of pediatric health expenditures
\citep{cohen2011cmc,simon2010cmc}, and randomized evidence to guide their
care is hard to generate because of heterogeneous comorbidities and
medical fragility. In the face of limited evidence, treatment decisions
skew toward high-intensity, high-risk options even when less intense
options are equally reasonable \citep{wennberg2002,wennberg2014}. Shared
decision-making (SDM) is a promising but understudied approach to align
decisions with family values instead of defaulting to high intensity
\citep{charles1997,lin2018sdmcmc}, and CMC specifically receive
lower-quality SDM than non-complex children \citep{lin2018sdmcmc}. Our own
clinical observations further motivated automated measurement in this
population.

Measuring SDM currently requires human coding of recorded encounters using
structured schemes such as the Clayman et al.\ coding system
\citep{clayman2012}, which is expensive, slow, and retrospective. This is the
bottleneck that keeps SDM measurement out of routine care and out of
feedback loops that could change clinician behavior, and it persists even
when efficacious SDM interventions exist \citep{stacey2024}. Recent work
suggests large language models (LLMs) may help approximate human SDM ratings
from transcripts \citep{waddell2024,oh2024,pandiselvaraj2024}, raising the
question of whether prompting alone is sufficient.

We identify three unexamined assumptions in that reading. \textbf{Setting}:
much of this prior work uses simulated, scripted, or narrowly scoped
encounters, while real pediatric surgical consultations are fragmented,
multi-party, and full of backchannel. \textbf{Privacy}: identifiable clinical
audio and transcripts require approved safeguards, motivating local inference.
\textbf{Evaluation}: small-$n$ clinical corpora are especially vulnerable to
optimistic estimates when recordings, patients, and labeled prompt exemplars
are not separated consistently.

This paper makes three contributions: (i) a head-to-head comparison of
zero-shot prompting, supervised learning, and their hybrid on real
encounters, entirely on-premises; (ii) an audit of corpus-specific leakage
paths involving repeated patients and labeled prompt exemplars; and (iii) a
descriptive patient-grouped learning curve for the supervised component.

\section{Objectives}

To determine whether zero-shot LLM prompting is sufficient to detect SDM
behaviors in real clinical encounters and whether supervised learning adds
value under patient-grouped, nested evaluation.

\section{Materials and Methods}

\subsection{Setting and data}

Twenty-one audio-recorded outpatient encounters in which families of
children with neuromuscular scoliosis and their surgical providers
discussed operative management. Recordings come from \textbf{19 unique
patients}: two patients contributed two recordings each. Audio was
transcribed and diarized locally (Whisper \citep{radford2022whisper} +
pyannote.audio \citep{bredin2020pyannote}), producing \textbf{7{,}566
speaker-attributed utterance segments} across approximately 6.1 hours.

Treating the 21 recordings as 21 independent units is incorrect and is one
of the leakage sources audited in Section~\ref{sec:leakage}.

\subsection{Labels}

Trained coders labeled excerpts in Dedoose against a 12-behavior SDM
codebook (Table~\ref{tab:codebook}). Labels are multi-label: a segment may
carry zero or several behaviors, and behaviors are coded as contiguous spans
rather than isolated utterances. For model training and evaluation, all
available coder labels were aligned to transcript segments and merged by
union. Twenty recordings had labels from both principal coders; one had labels
from one principal coder.

Human--human agreement on the 18 original double-coded recordings was
\textbf{macro Cohen's $\kappa = 0.695$} (per-behavior range 0.41--0.83).
This is an estimate of how reproducibly two people applied the codebook. It
is contextual rather than a model ceiling because the model target is the
union of available labels and the model and human comparisons use different
reference constructions.

\begin{table}[htbp]
\centering
\caption{SDM behaviors and human agreement on the 18 original double-coded
recordings. Segment counts are shown separately for the two coders.}
\label{tab:codebook}
\begin{tabular}{clrrr}
\toprule
\# & Behavior & Human $\kappa$ & Coder 1 & Coder 2 \\
\midrule
0  & rationale for option              & 0.692 & 1{,}572 & 1{,}458 \\
1  & patient outcome expectations      & 0.785 & 625 & 572 \\
2  & definition of option              & 0.415 & 175 & 191 \\
3  & risks/cons                        & 0.740 & 631 & 534 \\
4  & benefits/pros                     & 0.751 & 132 & 143 \\
5  & patient preferences \& values     & 0.662 & 441 & 524 \\
6  & patient understanding confirmed   & 0.605 & 1{,}642 & 2{,}074 \\
7  & plan for follow-up                & 0.833 & 765 & 885 \\
8  & provider preferences \& values    & 0.628 & 1{,}430 & 1{,}639 \\
9  & degree of decision sharing        & 0.803 & 108 & 125 \\
10 & process or procedure              & 0.668 & 998 & 1{,}390 \\
11 & patient self-efficacy             & 0.764 & 34 & 47 \\
\midrule
   & \textbf{Macro}                    & \textbf{0.695} & & \\
\bottomrule
\end{tabular}
\end{table}

\subsection{Approaches compared}

All three operate on the same 7-utterance sliding window centered on the
segment being classified, with speaker roles marked inline.

\textbf{(A) Zero-shot LLM.} Qwen 2.5 32B \citep{yang2024qwen25} (4-bit),
served locally via Ollama. The prompt supplies the 12-behavior codebook and
brief sub-theme cues, and requests a JSON object mapping each present
behavior to a graded confidence in $[0,1]$. Graded output (rather than
binary yes/no) gives the downstream combiner a richer feature. Generation
used temperature 0.

\textbf{(B) Supervised.} Frozen Qwen3-Embedding-0.6B
\citep{zhang2025qwen3embedding} representations of the window, concatenated
with speaker-role features, feeding twelve independent class-balanced
logistic regressions (one per behavior).
Each used $C=1.0$, class balancing, and the liblinear solver.

\textbf{(C) Hybrid.} For each behavior, a class-balanced logistic regression
combines the log-odds of the zero-shot LLM and supervised probabilities.
The stacker is fit only on inner patient-grouped predictions.

We separately audited a four-shot configuration with four deterministic,
rare-topic-first labeled exemplars. Although each query excluded exemplars
from its own patient, the precomputed cache was not fully nested with respect
to the outer folds; those estimates are therefore not used as primary results.

Everything runs on-premises. No transcript, audio, or derived text leaves
the institution at any stage.

\subsection{Evaluation protocol}
\label{sec:evaluation}

This is treated as a first-class component of the method, not boilerplate.

\begin{enumerate}
  \item \textbf{Patient-grouped outer folds.} Each outer fold holds out
  every recording belonging to one patient (19 folds).
  \item \textbf{Inner cross-fitting.} Within each outer training set,
  5-fold patient-grouped cross-fitting produces out-of-fold supervised
  probabilities. Combiner weights and per-behavior decision thresholds are
  fit on those out-of-fold probabilities only, never on in-sample supervised
  predictions. Thresholds are selected from 0.05 to 0.95 in increments of
  0.05 by maximizing training-fold $\kappa$.
  \item \textbf{Cluster bootstrap.} 95\% CIs resample whole patients
  ($n=1{,}000$), not utterances or recordings, because segments within a
  patient are not independent.
\end{enumerate}

Primary metric is macro Cohen's $\kappa$ \citep{cohen1960} across the 12
behaviors, computed after pooling all outer-fold predictions and averaged
over behaviors that are not structurally undefined. Marginal intervals use
1{,}000 patient-cluster bootstrap resamples; paired model differences use
5{,}000 resamples of the same patient clusters.

\section{Results}

\subsection{Zero-shot prompting alone is the weakest approach}

\begin{table}[htbp]
\centering
\caption{Performance under the patient-grouped nested protocol
(19 patient groups, 7{,}566 segments). Confidence intervals are marginal
patient-bootstrap intervals; the human--human value is contextual and not
directly comparable.}
\label{tab:main-results}
\begin{tabular}{lrrr}
\toprule
Approach & Macro $\kappa$ & 95\% CI & $\Delta$ vs.\ LLM \\
\midrule
Zero-shot LLM (Qwen 2.5 32B) & 0.139 & 0.111--0.164 & --- \\
\textbf{Supervised only} & \textbf{0.227} & \textbf{0.186--0.262} & \textbf{+0.088} \\
Hybrid, zero-shot logistic stack & 0.242 & 0.198--0.284 & +0.103 \\
\textit{Human--human reference} & \textit{0.695} & --- & --- \\
\bottomrule
\end{tabular}
\end{table}

The supervised model improved macro $\kappa$ over zero-shot prompting by
$0.088$; the 95\% paired patient-bootstrap interval was 0.051--0.119. The
zero-shot logistic stack had a higher point estimate than the supervised
model, but its marginal interval overlapped substantially; we do not claim a
paired improvement for that contrast.

Per-behavior point estimates (Table~\ref{tab:per-topic}) show why the macro
average matters. The zero-shot LLM was strongest for \textit{definition of
option}, while the supervised model was markedly stronger for behaviors such
as \textit{patient outcome expectations}, \textit{plan for follow-up}, and
\textit{provider preferences \& values}. These are descriptive contrasts; no
per-behavior uncertainty intervals were computed.

\begin{table}[htbp]
\centering
\caption{Per-behavior $\kappa$ point estimates under the primary zero-shot
protocol.}
\label{tab:per-topic}
\begin{tabular}{lrrr}
\toprule
Behavior & Zero-shot LLM & Supervised & Logistic stack \\
\midrule
rationale for option             & 0.093 & 0.274 & 0.281 \\
patient outcome expectations     & 0.061 & 0.380 & 0.388 \\
definition of option             & 0.302 & 0.198 & 0.293 \\
risks/cons                       & 0.404 & 0.448 & 0.428 \\
benefits/pros                    & 0.290 & 0.255 & 0.370 \\
patient preferences \& values    & 0.101 & 0.121 & 0.140 \\
patient understanding confirmed  & 0.007 & 0.098 & 0.119 \\
plan for follow-up               & 0.146 & 0.337 & 0.338 \\
provider preferences \& values   & 0.008 & 0.246 & 0.249 \\
degree of decision sharing       & 0.207 & 0.205 & 0.132 \\
process or procedure             & 0.055 & 0.191 & 0.182 \\
patient self-efficacy            & 0.000 & $-0.023$ & $-0.014$ \\
\bottomrule
\end{tabular}
\end{table}

\textbf{An honest failure.} \textit{patient self-efficacy} is not reliably
detected by any approach, despite high human agreement ($\kappa=0.764$,
Table~\ref{tab:codebook}). It has 126 labeled segments but only
\textbf{10 contiguous spans across 6 patients} in the full union reference:
the segment count badly
overstates the number of independent examples. We report this as
insufficient independent positive events for stable model assessment, not
as evidence the behavior is inherently undetectable; it needs targeted data
collection before another modeling attempt is informative.

\subsection{A leakage audit}
\label{sec:leakage}

We found three ways in which apparently patient-aware evaluation could still
allow label information to cross an evaluation boundary. Because these issues
interact and the affected pipelines do not expose identical predictions, we do
not assign a numerical effect to each one.

\begin{itemize}
  \item \textbf{Sibling recordings.} Two patients contributed two
  recordings each. Under recording-level folds, one recording of a patient
  trains a model evaluated on the other.
  \item \textbf{Exemplar leakage.} Few-shot exemplars selected by excluding
  the held-out recording still admitted a sibling recording of the same
  patient. Replaying the deterministic selection showed this fired for 3
  of 21 recordings.
  \item \textbf{Outer-fold leakage through a precomputed prompt cache.}
  Excluding the patient being predicted is not sufficient when those
  predictions are later used to fit a stacker. For three of 19 outer folds,
  labeled exemplars from the outer held-out patient had entered cached
  predictions for patients in the outer training set. The cache was therefore
  valid for leave-one-patient-out standalone prompting but not for nested
  downstream fitting. We exclude the affected few-shot stack and metadata-gate
  estimates from the primary results.
\end{itemize}

These failures are easy to miss because the final test prediction can exclude
its own patient while the training features used by a downstream model still
contain information from that patient. Future evaluations should generate
labeled prompt exemplars inside each outer fold and should report patient counts
alongside recording counts.

\subsection{Descriptive supervised learning curve}

\begin{figure}[htbp]
\centering
\includegraphics[width=0.78\textwidth]{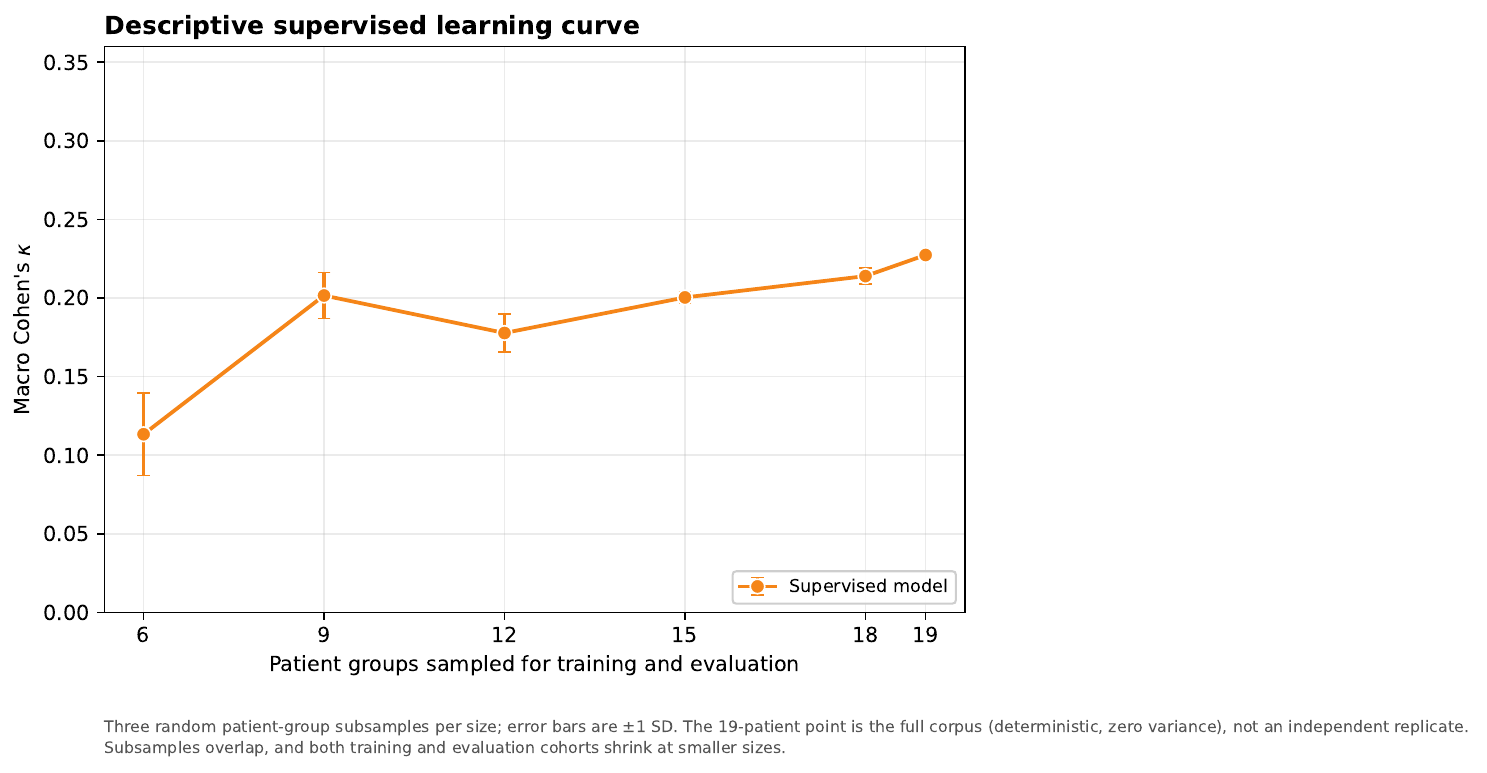}
\caption{Supervised-model macro $\kappa$ versus number of coded patients,
using three random patient-group subsamples per size. Error bars show
$\pm 1$ SD; 19 patients is the full corpus and therefore has zero variance.
The curve is descriptive and is not an extrapolation beyond the observed
sample.}
\label{fig:learning-curve}
\end{figure}

We subsampled patient groups (not recordings) at 6, 9, 12, 15, and 18
patients, repeated each size three times, and refit the supervised branch
under the nested protocol in Section~\ref{sec:evaluation}. Mean macro
$\kappa$ was 0.113, 0.202, 0.178, 0.200, and 0.214, respectively; the full
19-patient estimate was 0.227 (Figure~\ref{fig:learning-curve}). Across the
15 subsampled runs, Spearman's $\rho$ between patient count and $\kappa$ was
0.70.

This positive descriptive association suggests that additional labeled data
may improve the supervised model, but it does not establish a learning-curve
shape or absence of saturation. There were only three replicates per size,
subsamples overlap, training and evaluation cohorts shrink together, and
patient count is a coarse proxy for data volume: the 6-patient subsets ranged
from 909 to 3{,}358 segments.

\section{Discussion}

\textbf{Why zero-shot prompting may underperform here.} The behavior-level
pattern suggests two hypotheses for future testing. Several SDM behaviors are
defined by interactional structure (for example, a clinician checks
understanding and a family member responds) rather than by topic words alone.
In addition, span labels reflect clinic-specific realizations that a general
model has not observed. The present study does not isolate these mechanisms.

\textbf{What this does and does not say about LLMs.} It is not a claim that
LLMs are unsuited to this task. It is a claim that zero-shot prompting of a
general model was insufficient in this dataset and configuration, and that a
small supervised model over frozen representations was a stronger baseline.
The study evaluates one local LLM, one zero-shot prompt family, one codebook,
and one clinical population; it does not establish a general ranking of LLM
and supervised approaches.

\textbf{Evaluation is part of the model.} Grouping by patient is necessary but
not sufficient for few-shot pipelines. Every labeled artifact used to create
training features, including prompt exemplars and cached LLM outputs, must be
generated inside the appropriate outer fold. This requirement becomes easy to
violate when expensive model outputs are cached once and reused.

\textbf{Limitations.} This was a single-site study of 19 patients with no
external test set; all configurations were developed using this corpus, so the
intervals do not capture uncertainty from prior model and prompt selection.
The union reference favors sensitivity and is not an adjudicated consensus.
Per-behavior estimates lack uncertainty intervals. Offline evaluation uses
symmetric future context and does not establish streaming or clinical
performance. The descriptive learning curve uses only three overlapping
subsamples per size. Finally, the results apply to one quantized LLM and one
prompt; external and prospective evaluation are required.

\section{Conclusion}

In these real pediatric encounters, a small supervised model agreed with the
SDM reference labels more strongly than zero-shot Qwen 2.5 32B under paired,
patient-grouped evaluation. The result supports supervised learning as a
necessary baseline for clinical-dialogue measurement and shows that leakage
control must extend to labeled prompt exemplars and prediction caches. Larger,
externally validated cohorts are needed before drawing conclusions about
clinical use or about LLMs beyond this model and prompt.

\end{document}